\pdfoutput=1

\documentclass[11pt,a4paper]{article}
\usepackage{authblk}
\usepackage[hyperref]{eacl2021}
\usepackage{times}

\usepackage{latexsym}
\usepackage{pifont}
\usepackage{comment}
\usepackage{multirow}
\usepackage{float}
\restylefloat{table}
\usepackage{tikz}
\usetikzlibrary{shapes.geometric, arrows}
\usepackage{microtype}

\aclfinalcopy % Uncomment this line for the final submission
\title{Hopeful\_Men@LT-EDI-EACL2021: Hope Speech Detection Using Indic Transliteration and Transformers}
\author[1]{\textbf{Ishan Sanjeev Upadhyay*}}
\author[1]{\textbf{Nikhil E*}}
\author[2]{\textbf{Anshul Wadhawan}}
\author[3]{\textbf{Radhika Mamidi}}
\affil[1,3]{International Institute of Information Technology, Hyderabad}
\affil[2]{Flipkart Private Limited}
\affil[1]{\texttt{\{ishan.sanjeev, nikhil.e\}@research.iiit.ac.in} }
\affil[2]{\texttt{anshul.wadhwan@flipkart.com}}
\affil[3]{\texttt{radhika.mamidi@iiit.ac.in}}
\date{}

\begin{document}
\maketitle
\begin{abstract}
This paper aims to describe the approach we used to detect hope speech in the HopeEDI dataset. We experimented with two approaches. In the first approach, we used contextual embeddings to train classifiers using logistic regression, random forest, SVM, and LSTM based models.The second approach involved using a majority voting ensemble of 11 models which were obtained by fine-tuning pre-trained transformer models (BERT, ALBERT, RoBERTa, IndicBERT) after adding an output layer. We found that the second approach was superior for English, Tamil and Malayalam. Our solution got a weighted F1 score of 0.93, 0.75 and 0.49 for English,Malayalam and Tamil respectively. Our solution ranked first in English, eighth in Malayalam and eleventh in Tamil.
\end{abstract}
\let\thefootnote\relax\footnotetext{*These authors contributed equally to this work}
\section{Introduction}

The spread of hate speech on social media is a problem that still exists today. While there have been attempts made at hate speech detection \citep{schmidt,lee} to stop the spread of negativity, this form of censorship can also be misused to obstruct rights and freedom of speech. Furthermore, hate speech tends to spread faster than non-hate speech \citep{Mathew}.While there has been a growing amount of marginalized people looking for support online \citep{Gowen,wang-jurgens-2018-going}, there has been a substantial amount of hate towards them too \citep{Mondal}. Therefore, detecting and promoting content that reduces hostility and increases hope is important. Hope speech detection can be seen as a rare positive mining task because hope speech constitutes a low percentage of overall content \citep{palakodety2020hope}. There has been work done on hope speech or help speech detection before that has used logistic regression and active learning techniques \citep{palakodety2020hope,Palakodety_KhudaBukhsh_Carbonell_Palakodety_KhudaBukhsh_Carbonell_2020}. In our paper, we will be doing the hope speech detection task on the HopeEDI dataset \citep{chakravarthi-2020-hopeedi} which consists of user comments from Youtube in English, Tamil and Malayalam. 

In this paper, we will first look at the task definition, followed by the methodology used. We will then look at the experiments and results followed by conclusion and future work. 
%\footnote{*These authors contributed equally to this work.}
\section{Task Definition}
The given problem is a comment level classification task for the identification of "hope speech" within YouTube comments, wherein they are to be classified as "Hope speech", "Not hope speech" and "Not in intended language". The data provided in the task was annotated at a per-comment basis wherein a comment could be composed of more than one sentence.

\section{Methodology}

%\pagebreak
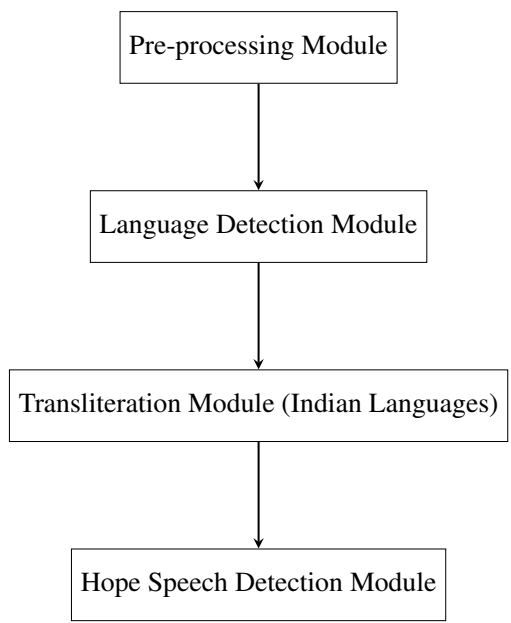
\begin{figure}
\centering
\scalebox{0.95}{%
\tikzstyle{startstop} = [rectangle,minimum width=3cm, minimum height=1cm,text centered, draw=black]
\tikzstyle{arrow} = [thick,->,>=stealth]
\begin{tikzpicture}[node distance=2cm]
\node (preprocessing) [startstop] {Pre-processing Module};

\node (lang) [startstop,below of=preprocessing, yshift=-0.5cm]  {Language Detection Module};

\node (translit) [startstop,below of=lang, yshift=-0.5cm] {Transliteration Module (Indian Languages)};

\node (hope) [startstop,below of=translit, yshift=-0.5cm] {Hope Speech Detection Module};

\draw [arrow] (preprocessing) -- (lang);
\draw [arrow] (lang) -- (translit);
\draw [arrow] (translit) -- (hope);

\end{tikzpicture}}
\caption{\label{pipeline} Methodology Pipeline }
\end{figure}

This section talks about the methodology that we have used to solve the task. As shown in Figure 1, the pipeline involves preprocessing, language detection, transliteration (for Indian languages), and hope speech detection. These steps are described in this section.

\subsection{Pre-processing Module}
The preprocessing module involved the following:
\begin{itemize}
  \item Removing special characters and excess whitespaces
  \item Removing emojis
  \item Make text lowercase. 
\end{itemize}
These steps were taken to make the text more uniform. Special characters like “@” and “\#” were removed because they did not serve as good features for classification and language detection. Emojis were removed because they were sparsely used in the dataset.

\subsection{Language Detection Module}

The task involves classifying text into hope, not-hope and not-language. Language detection module marks the not-language sentences. We use Google’s language detection library \citep{nakatani2010langdetect} to do this. The Tamil and Malayalam datasets are code-mixed. Inter-sentential,intra-sentential, tag code-mixing and code-mixing between Latin and native script is observed in the Tamil and Malayalam datasets. Google’s language detection library does not work on such code-mixed data. Since the Tamil and Malayalam sentences involve code-mixed data, language detection can not be done on them using the Google language detection library. We observed that sentences that were marked as not-Tamil and not-Malayalam were mostly English sentences with some of them being Hindi and other languages. Hence, we adopted a heuristic where we marked sentences as not-Tamil or not-Malayalam if the sentences were detected to be in English or Hindi, other sentences were assumed to belong to the respective language.

\subsection{Transliteration Module}
After language detection, sentences that are classified to be in Tamil and Malayalam undergo transliteration. Tamil and Malayalam text have code-mixing between Latin and native script, hence transliteration is done to make the entire text in the native script. This step is also important because it makes the text closer to the kind of text IndicBert is trained on. Transliteration was done by using the indic-transliteration library  \footnote{\url{https://github.com/sanskrit-coders/indic_transliteration}}.

\subsection{Hope Speech Detection Module}
After preprocessing and transliteration (for Indian languages), the text is sent to the hope speech detection module. The hope speech detection module is responsible for predicting if a text is hope speech or not hope speech. We have used the following for our experiment.

\subsubsection{Models}
\textbf{Transformers} \citep{vaswani} have performed well in various natural language processing (NLP) tasks. Unlike recurrent neural networks (RNN), transformers are non-sequential (ie. sentences are processed as a whole rather than word by word) and use self-attention at each input time step. Hence, they do not suffer from long dependency issues. Query, Key and Value are three different ways in which input vectors are used in the self-attention mechanism. The attention score for every input vector is calculated using a compatibility function which takes as input the query vector and all the keys. The final output is a weighted sum of values where the weights are the attention scores calculated by the compatibility function. We have used the following transformers in our experiment. We chose RoBERTa for our final model in English and ALBERT (IndicBERT) for Tamil and Malayalam. \par
BERT \citep{devlin-etal-2019-bert} is based on the transformer architecture. Using its multi-layer encode module,It is able to jointly utilize both left and right contexts across all layers to pre-train its bidirectional representations. BERT is trained on two unsupervised prediction tasks, next sentence prediction and masked language modelling. We have fine tuned  “bert-base-uncased”  model on the dataset for one of our experiments.
    
RoBERTa \citep{liu2019roberta} is a transformer architecture which is based on optimizations made to the BERT approach. It trains on more data and bigger batches, removes next sentence prediction objective that BERT used, trains on longer sequences and introduces dynamic masking (ie. mask tokens change during training epochs). RoBERTa outperforms BERT and XLNet on the GLUE benchmark. For the hope speech classification task in English, we fine-tuned the “roberta-base” model on the provided data. The roberta-base model is trained on 160 GB of English text from five different datasets.

ALBERT \citep{lan2020albert} is a transformer architecture based on BERT but with fewer parameters. There are two key changes made to ALBERT. The first is factorized embeddings parameterization, which decomposes the large vocabulary embedding matrix into smaller matrices. This makes it easier to grow hidden size without increasing the parameter size of vocabulary embeddings.  This step leads to a reduction in parameters by 80\% compared to BERT. The second is cross-layer parameter sharing, which prevents the parameters from increasing as the depth of the network increases. We fine-tuned the IndicBERT model for the hope speech classification task in Tamil and Malayalam. We used IndicBERT \citep{kakwani} which is a multilingual ALBERT model pre-trained on 12 major Indian languages. We also fine-tuned  "albert-base-v2" model for our experiment in English.

\textbf{LSTM}
Long Short-Term Memory \citep{Hochreiter} networks seek to solve the short-term memory or vanishing gradient problem that RNNs face. They do so by having internal gates that regulate the flow of information. Information flows through a mechanism known as cell states. The cell can make decisions about what to store, what to forget and what the next hidden state should be. This is done through internal mechanisms called gates which contain sigmoid activations.

\textbf{Random Forest Classifier}
Random forests \citep{Breiman} use an ensemble of a large number of decision trees generally trained with the bagging method. These decision trees are created using random subsamples of the given dataset with replacement (bootstrap dataset) and a random subset of the features. New samples are classified by choosing the prediction made by most decision trees (majority voting).

\textbf{Support Vector Machine}
Support vector machine (SVM) \citep{Hearst} is a supervised learning method that can be used for classification or regression. We have used SVM for classification. The objective of the SVM classification algorithm is to find the hyper-plane that most accurately differentiates two classes that have been plotted on a f dimensional plane where f is the number of features.

\textbf{Logistic Regression}
Logistic regression \citep{mccullagh}is a statistical model used for binary classification. It does so by using a logistic function to model the binary outcome. It can be extended for multiclass classification problems. 

\subsubsection{Ensemble Process}
Ensembles can help make better predictions by reducing the spread of predictions. Hence, lowering variance and improving accuracy. We used a voting based ensemble method where we trained N models on N different training and validation data obtained by random shuffling. We then chose the majority voting as the merging technique to produce our final prediction y. In majority voting, the final prediction y is decided based on which prediction is made by the majority of the models . We made two ensembles, one each of 7 models and 11 models and chose the ensemble that gave the best weighted F1 score.

\section{Experiments}

Initially, the entire database is preprocessed to remove extra tab spaces, punctuations, emojis, mentions and links. In the case of Malayalam and Tamil, we also transliterate the entire database. Then we distributed our experimentation procedure into two different approaches. In the first approach, we finetune our pre-trained masked language models using the train and validation splits for the purpose of making them more suitable to the subsequent classification task. Thereafter, contextual embeddings for each sentence in the dataset are produced by calculating the average of the second to last hidden layer for every single token in the sentence. We then trained Logistic Regression, Random Forest, SVM and RNN-based classifier models using these embeddings. In the second approach, all the sentences are encoded into tokens using the respective tokenizers and then we add a linear layer on top of the pre-trained model layers after dropout. All the layers of the devised model are then trained such that the error is back propagated through the entire architecture and the pre-trained weights of the model are modified to reflect the new database. For both these approaches, we then calculated predictions for the test split and reported performance metrics. For English, we try out three different pre-trained models: "roberta-base", "bert-base-uncased", and "albert-base-v2" for both the approaches. For Tamil and Malayalam however, only the IndicBERT model is applicable for either approach.

\begin{table}[H]
\scalebox{0.87}{%
\begin{tabular}{|p{0.1\textwidth}|p{0.09\textwidth}|l|p{0.1\textwidth}|p{0.06\textwidth}|} 
\hline
\textbf{Language} & \textbf{Database} & \textbf{Hope} & \textbf{Not Hope} & \textbf{Other Lang.} \\ 
\hline
English & Train & 1962 & 20778 & 22 \\ 
\cline{2-5} 
& Dev & 242 & 2569 & 2 \\ 
\hline
Tamil & Train & 6327 & 7872 & 1961 \\
\cline{2-5}
& Dev & 757 & 998 & 263 \\ 
\hline
Malayalam & Train & 1668 & 6205 & 691 \\
\cline{2-5}
& Dev & 190 & 784 & 96 \\ 
\hline

\end{tabular}}
\caption{\label{trainDev} Data distribution by class}
\end{table}

\begin{table}[H]
\centering
\scalebox{0.87}{%
\begin{tabular}{|l|l|l|l|}
\hline
     & \textbf{English} & \textbf{Tamil} & \textbf{Malayalam}  \\
\hline
Training    & 22762   & 16160 & 2564       \\
\hline
Development & 2843    & 2018  & 1070       \\
\hline
Test        & 2846    & 2020  & 1071       \\
\hline
Total       & 28451   & 20198 & 10705     \\
\hline
\end{tabular}}
\caption{\label{trainDev} Data distribution by language}
\end{table}

\subsection{Dataset}
The HopeEDI dataset consists of Youtube comments marked as “hope”, “not hope” and “other language” in three languages: English, Tamil and Malayalam. The distribution of hope, not hope and other language tag in the training and development datasets is shown in table 1. The ratio of hope to not hope is around 0.09 in English, 0.26 in Malayalam and 0.79 in Tamil. Table 2 shows the data distribution between training, development and test datasets. There are a total of 28,451 comments in English, 10,705 comments in Malayalam and 20,198 comments in Tamil. Data in Tamil and Telugu has code-mixing. In the English dataset, there are instances where English comments are annotated as not English. For example, “Fox News is pure Garbage!” is annotated as not English in the training set. This contributes some noise to the English dataset.

\begin{table*}[]
\centering
\begin{tabular}{|l|l|l|l|l|l|l|l|}
\hline
\textbf{Model} & \textbf{Method} & \textbf{Macro} & \textbf{Weighted} & \textbf{Macro} & \textbf{Weighted} & \textbf{Macro} & \textbf{Weighted}\\ 
 & \textbf{Used} & \textbf{Precision} & \textbf{Precision} & \textbf{Recall} & \textbf{Recall} & \textbf{F1-Score} & \textbf{F1-Score} \\ \hline
BERT & E + LR & 0.778 & 0.911 & 0.656 & 0.924 & 0.695 & 0.913 \\ \cline{2-8}
 & E + RF & 0.834 & 0.902 & 0.526 & 0.916 & 0.528 & 0.881 \\ \cline{2-8}
 & E + SVM & 0.771 & 0.86 & 0.489 & 0.866 & 0.488 & 0.837 \\ \cline{2-8}
& E + LSTM & 0.456 & 0.833 & 0.500 & 0.913 & 0.477 & 0.871 \\ \cline{2-8}
& FT & 0.759 & 0.915 & 0.728 & 0.915 & 0.742 & 0.915 \\ \hline
ALBERT & E + LR & 0.703 & 0.881 & 0.538 & 0.912 & 0.549 & 0.883 \\ \cline{2-8}
 & E + RF & 0.832 & 0.900 & 0.506 & 0.914 & 0.491 & 0.874 \\ \cline{2-8}
 & E + SVM & 0.456 & 0.833 & 0.500 & 0.913 & 0.477 & 0.871 \\ \cline{2-8}
& E + LSTM & 0.657 & 0.878 & 0.571 & 0.905 & 0.591 & 0.887 \\ \cline{2-8}
& FT & 0.755 & 0.916 & 0.705 & 0.924 & 0.725 & 0.919 \\ \hline
RoBERTa & E + LR & 0.794 & 0.914 & 0.657 & 0.926 & 0.700 & 0.915 \\ \cline{2-8}
 & E + RF & 0.840 & 0.905 & 0.535 & 0.917 & 0.544 & 0.885 \\ \cline{2-8}
& E + SVM & 0.821 & 0.899 & 0.517 & 0.915 & 0.512 & 0.878 \\ \cline{2-8}
& E + LSTM & 0.791 & 0.918 & 0.693 & 0.928 & 0.729 & 0.921 \\ \cline{2-8}
& FT & 0.753 & 0.915 & 0.748 & 0.922 & 0.745 & \textbf{0.923} \\ \hline
\end{tabular}
\caption{\label{results} Metrics for English language%\begin{center} \textbf{E-Contextualized Embeddings} \end{center} \\ \begin{center} \textbf{LR-Logistic Regression} \end{center} \\ \begin{center} \textbf{RF-Random Forest} \end{center} \\ \begin{center} \textbf{SVM-Support Vector Machines} \end{center} \\ \begin{center} \\ \textbf{RNN-LSTM based classifier} \end{center} \\ \begin{center} \textbf{FT-Finetuned model with output layer} \end{center}
}
\end{table*}

\begin{table*}[]
\centering
\begin{tabular}{|l|l|l|l|l|l|l|l|}
\hline
\textbf{Model} & \textbf{Method} & \textbf{Macro} & \textbf{Weighted} & \textbf{Macro} & \textbf{Weighted} & \textbf{Macro} & \textbf{Weighted}\\ 
 & \textbf{Used} & \textbf{Precision} & \textbf{Precision} & \textbf{Recall} & \textbf{Recall} & \textbf{F1-Score} & \textbf{F1-Score} \\ \hline
Indic & E + LR & 0.473 & 0.482 & 0.484 & 0.520 & 0.441 & 0.464 \\ \cline{2-8}
-BERT & E + RF & 0.511 & 0.516 & 0.506 & 0.544 & 0.458 & 0.482 \\ \cline{2-8}
& E + SVM & 0.278 & 0.309 & 0.500 & 0.556 & 0.357 & 0.397 \\ \cline{2-8}
& E + LSTM & 0.591 & 0.587 & 0.501 & 0.557 & 0.364 & 0.403 \\ \cline{2-8}
& FT & 0.635 & 0.637 & 0.627 & 0.636 & 0.623 & \textbf{0.629} \\ \hline
\end{tabular}
\caption{\label{results} Metrics for Tamil language%\begin{center} \textbf{E-Contextualized Embeddings} \end{center} \\ \begin{center} \textbf{LR-Logistic Regression} \end{center} \\ \begin{center} \textbf{RF-Random Forest} \end{center} \\ \begin{center} \textbf{SVM-Support Vector Machines} \end{center} \\ \begin{center} \\ \textbf{RNN-LSTM based classifier} \end{center} \\ \begin{center} \textbf{FT-Finetuned model with output layer} \end{center}
}
\end{table*}

\begin{table*}[]
\centering
\begin{tabular}{|l|l|l|l|l|l|l|l|}
\hline
\textbf{Model} & \textbf{Method} & \textbf{Macro} & \textbf{Weighted} & \textbf{Macro} & \textbf{Weighted} & \textbf{Macro} & \textbf{Weighted}\\ 
 & \textbf{Used} & \textbf{Precision} & \textbf{Precision} & \textbf{Recall} & \textbf{Recall} & \textbf{F1-Score} & \textbf{F1-Score} \\ \hline
Indic & E + LR & 0.645 & 0.729 & 0.501 & 0.790 & 0.447 & 0.699 \\ \cline{2-8}
-BERT & E + RF & 0.395 & 0.623 & 0.499 & 0.788 & 0.440 & 0.696 \\ \cline{2-8}
 & E + SVM & 0.386 & 0.610 & 0.492 & 0.777 & 0.433 & 0.683 \\ \cline{2-8}
& E + LSTM & 0.367 & 0.579 & 0.471 & 0.745 & 0.413 & 0.652 \\ \cline{2-8}
& FT & 0.776 & 0.842 & 0.743 & 0.842 & 0.756 & \textbf{0.837} \\ \hline
\end{tabular}
\caption{\label{results} Metrics for Malayalam language%\begin{center} \textbf{E-Contextualized Embeddings} \end{center} \\ \begin{center} \textbf{LR-Logistic Regression} \end{center} \\ \begin{center} \textbf{RF-Random Forest} \end{center} \\ \begin{center} \textbf{SVM-Support Vector Machines} \end{center} \\ \begin{center} \\ \textbf{RNN-LSTM based classifier} \end{center} \\ \begin{center} \textbf{FT-Finetuned model with output layer} \end{center}
}
\end{table*}

\subsection {System Settings}
In the first approach, we run the task of masked language modelling on our database for 4 epochs for each of the 5 model-database combinations. Afterwards, the sentence input token length is limited to 512 and the embeddings extracted by evaluation on the input sequences by the model are of length 768. The RNN based classifier is composed of an LSTM layer and two dense layers. In the second approach, the encoded sentences are in the form of a data loader class, containing the respective input IDs and attention masks, with a batch size of 16. These are then passed into a model that implements a dropout of 30\% and the output from the final linear layer is used for classification.

\subsection{Evaluation Metrics}
We used F1 and weighted F1 scores for evaluating our model. 
\begin{center}
\begin{math} F1\: Score = 2 \times \frac{  (precision \times recall) }{(precision+recall)}
\end{math}
\end{center}
Weighted F1 scores are calculated by taking the F1 scores for each label and then doing a weighted average by the number of true instances of each label.
\begin{center}
\begin{math} weighted\: F1 = \frac{  (F1_{i} \times y_{i} + F1_{j} \times y_{j} ) }{(y_{i}+y_{j})}
\end{math}
\end{center}
$y_{i}$ and $y_{j}$ are the number of true instances of class $i$ and class $j$ respectively and $F1_{i}$  and $F1_{j}$ are the F1 scores of class $i$ and $j$ respectively.

\subsection{Results}
Our experimentation involved two approaches. In the first approach, we used contextual embeddings (E) to train classifiers using logistic regression (LR), random forest (RF), SVM, and LSTM based models. In the second approach we used an ensemble of 11 models which were generated by fine-tuning (FT) pre-trained transformer models after adding an output layer. We used majority voting to get our final prediction.  Results for both the approaches on our test split generated from the provided train and dev datasets are reported in Tables 3, 4 and 5 for English, Tamil and Malayalam respectively. We report the macro-averaged and weighted recall, precision and F1-score for each possible model-method combination. While the weighted F1 scores are more representative of how well a model performs, the disparity between the weighted and macro-averaged scores demonstrates how disproportionate a certain model’s effectiveness is in predicting the different classes. For English, the second approach involving finetuning is the best performing one for each of the models tested, closely followed by the Logistic Regression and LSTM-based methods in the first approach. The roberta-base model seems to have a slight edge over the other two tested models. For Tamil and Malayalam, the second approach is still the best performer, but by a greater margin.

%\pagebreak

\section{Conclusion and Future Work}
In this paper, we presented our approach for hope speech detection in  English, Tamil, and Malayalam on the HopeEDI dataset. We used two approaches. The first approach involved using contextual embeddings to train various classifiers. The second approach involved using a majority voting ensemble of 11 models which were obtained by fine-tuning pre-trained transformer models. The second approach using the roberta-base model was the best performing model for English, giving a weighted F1 score of 0.93. The second approach using IndicBERT model gave the best performance for Tamil and Malayalam, giving a weighted F1 score of 0.75 for Malayalam and 0.49 for Tamil. In the future, we plan to fine-tune transformers pre-trained on code mixed data. Data augmentation methods like synonym replacement and random insertion could be used to fine-tune the model on more data. 

\bibliography{eacl2021}
\bibliographystyle{acl_natbib}

\end{document}